\documentclass[journal]{IEEEtran}

\usepackage{graphicx}
\usepackage{CJKutf8}
\usepackage{multirow}
\usepackage{array}
\usepackage{tabularx}
\usepackage{makecell}
\usepackage{amsmath} 
\usepackage{amssymb}
\usepackage{tablefootnote}
\usepackage{subcaption}
\usepackage{diagbox}
\usepackage{booktabs}
\usepackage{xcolor}
\usepackage{color, soul} %用color, 和 soul 包
\usepackage{colortbl}
\usepackage{makecell}
\usepackage{arydshln}
\usepackage{algorithmic}
\usepackage{algorithm}
\ifCLASSINFOpdf
\else
\fi

\begin{document}
\title{Head-driven Phrase Structure Parsing in O($n^3$) Time Complexity}

\author{Zuchao Li,
        Junru Zhou,
        Hai Zhao,
        and Kevin Parnow
        \thanks{This paper was partially supported by National Key Research and Development Program of China (No. 2017YFB0304100), Key Projects of National Natural Science Foundation of China (U1836222 and 61733011), Huawei-SJTU long term AI project, Cutting-edge Machine Reading Comprehension and Language Model (Corresponding author: Hai Zhao).}
        \thanks{Zuchao Li, Junru Zhou, Hai Zhao, and Kevin Parnow are with the Department
			of Computer Science and Engineering, Shanghai Jiao Tong University, 800 Dongchuan Road, Minhang District, Shanghai, China, 200240. (e-mail: charlee@sjtu.edu.cn; zhoujunru@sjtu.edu.cn; zhaohai@cs.sjtu.edu.cn; parnow@sjtu.edu.cn).}}


\markboth{Arxiv, May~2021}%
{Li \MakeLowercase{\textit{et al.}}: Head-driven Phrase Structure Parsing in O($n^3$) Time Complexity}



\maketitle

\begin{abstract}
% Performance and efficiency are two key objectives for syntactic parsing. 
% Constituent and dependency parsing, as the two classic forms of syntactic parsing, have been found to be advantageous in joint training and decoding under a uniform formalism, Head-driven Phrase Structure Grammar (HPSG). 
% However, in the joint decoding of such unified grammar parses, the time complexity $O(n^5)$ is higher than that of individual decoding since more factors have to be considered in the decoding process.
% In this paper, we revisit the head parameter in HPSG for explicitly considering that in joint parsing and reduce the expensive time cost which results in a novel performance-preserved parser in O(n$^3$) time complexity. 
% Evaluations on typical and multilingual syntactic parsing benchmarks verified our claims. 
% In addition, we also examined the sources of performance improvements in HPSG-based parsing and explored the scenario where only one single treebank is available with a stronger base during training, as well as the impact of conversions from constituency to dependency with different head rules, which presents a more effective, more in-depth and general HPSG-based parsing research.
Constituent and dependency parsing, the two classic forms of syntactic parsing, have been found to benefit from joint training and decoding under a uniform formalism, Head-driven Phrase Structure Grammar (HPSG). However, decoding this unified grammar has a higher time complexity ($O(n^5)$) than decoding either form individually ($O(n^3)$) since more factors have to be considered during decoding. We thus propose an improved head scorer that helps achieve a novel performance-preserved parser in $O$($n^3$) time complexity. 
Furthermore, on the basis of this proposed practical HPSG parser, we investigated the strengths of HPSG-based parsing and explored the general method of training an HPSG-based parser from only a constituent or dependency annotations in a multilingual scenario. We thus present a more effective, more in-depth, and general work on HPSG parsing.
\end{abstract}


% \fi
\IEEEpeerreviewmaketitle

\section{Introduction}

\begin{figure}
    \centering
    \includegraphics[width=0.5\textwidth]{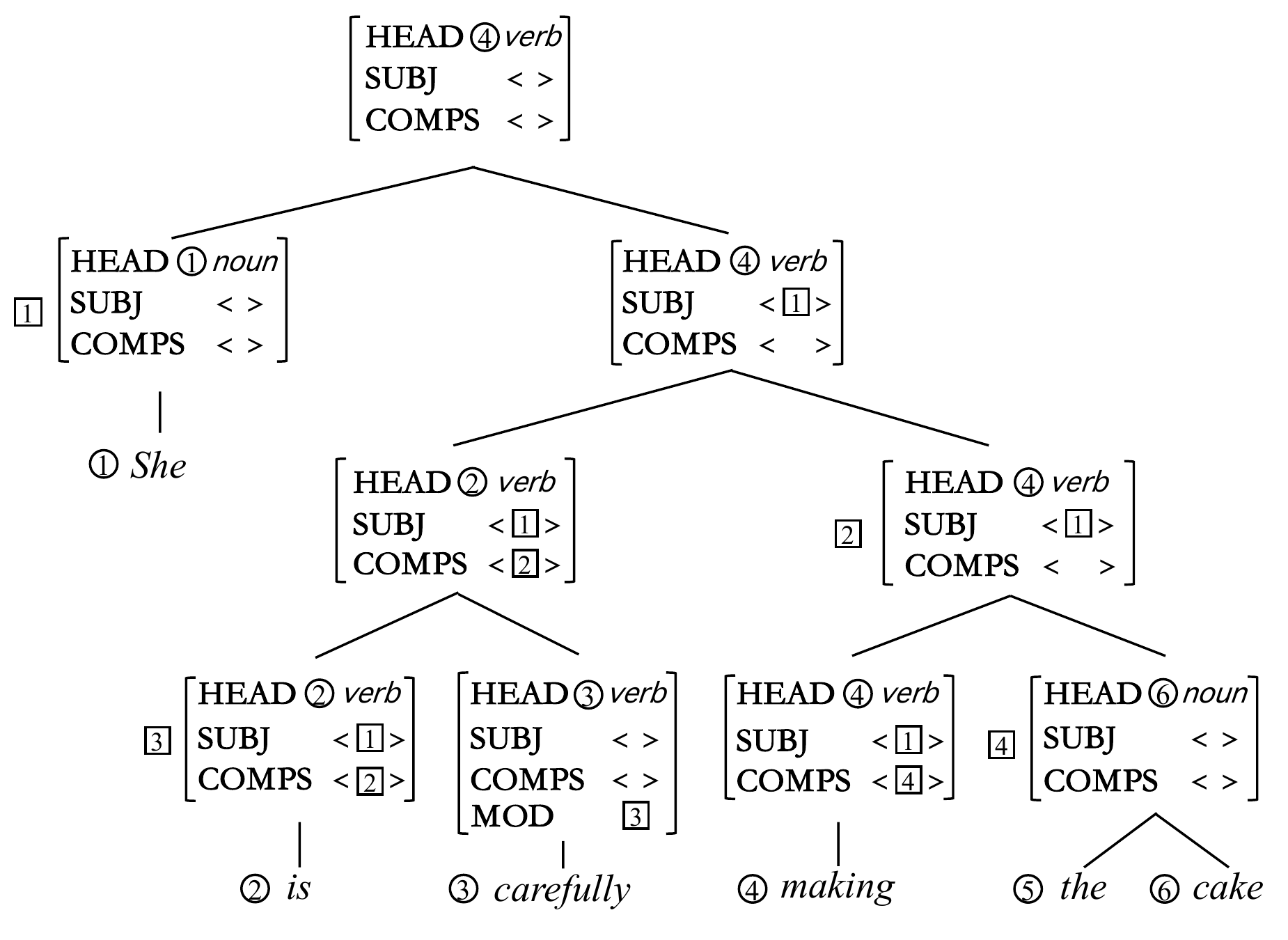}
    \caption{A HPSG tree for sentence ``\textit{She is carefully making the cake}." The numbers in circles after HEAD indicate the id of HPSG head words.}
    \label{fig:hpsg}
\end{figure}

% HPSG
%As a scientific study of human languages, linguistic grammar theory is concerned with revealing the nature of the mental grammar which represents speaker’s knowledge of their language \cite{hayes2013linguistics}. 
\IEEEPARstart{H}{ead-driven} Phrase Structure Grammar (HPSG) \cite{pollard1994head}, Generative Grammar \cite{chomsky1988generative},  Tree Adjoining Grammar (TAG) \cite{joshi1975tree}, Lexical Function Grammar (LFG) \cite{kaplan1982lexical}, and Combinatory Categorial Grammar (CCG) \cite{steedman2000syntactic} are known as the most sophisticated grammar frameworks for syntactic phrase structures.
HPSG, a lexicalized grammar, explains linguistic phenomena elegantly with a small number of grammar rules (phrase structure rules and lexical rules) and a number of complex lexical entries. In HPSG parse trees (shown in Figure \ref{fig:hpsg}), language symbols are divided into categories. Attribute Value Matrices (AVMs) that include phonological, syntactic, and semantic properties are the mechanisms for categorizing symbols.

% Head Parameter in HPSG
Strictly, HPSG is neither entirely a dependency grammar nor a phrase structure grammar.  It can be considered a Generative Grammar in the sense that it is explicit and formalized \cite{muller2019hpsg}, and it is also very similar to Categorial Grammar \cite{muller2010persian,kubota2020hpsg}. Alternatively, it can be treated as a synthesis of multiple grammar theories.
Works on parametric approaches to interlinguistic variation and grammar universals have drawn a great deal of attention \cite{hoeksema1992head}. One of the most basic cases of language variation is the position of the head element within a phrase (constituent). This illustrates the importance of the head parameter.
The constituent and head in HPSG follow the Head Feature Principle (HFP), that is, ``\textit{the head value of any headed phrase is structure-shared with the HEAD value of the head daughter. The effect of the HFP is to guarantee that headed phrases really are projections of their head daughter}." Because of this, using the schema of HPSG to develop a parser that combines constituency and dependency structures is possible.

% Zhou & Zhao HPSG Parsing
\cite{zhou-zhao-2019-head} made the first attempt to exploit strengths of the dependency and constituency representation formalisms and combine them as HPSG while using joint learning and decoding.
On the one hand, the joint learning and decoding improves parsing performance. On the other hand, this unification of grammars makes training a single model to produce two forms of parse trees and thus accommodate diverse downstream tasks possible, which reduces the cost of model training.
Despite these performance gains, there are still some drawbacks for this work.
In \cite{zhou-zhao-2019-head}, the training for constituent and dependency syntactic structures is performed jointly and is similar to the general joint training of constituent and dependency parsing models. 
The main innovation lies in the \textit{joint span} structure and the joint decoding algorithm based on this proposed structure. 
In this decoding algorithm, when considering constituent (span) partitioning, the dependency arcs are also considered as constraints by following the HFP. Thus the overall time complexity is expensive, and it actually reaches $O(n^5)$.

% ours
To address this inefficiency in neural HPSG-based constituency and dependency parsing, we present a novel framework that combines strengths from the approach proposed in \cite{zhou-zhao-2019-head} and features of the head parameter in HPSG parses to decrease the decoding time cost while retaining HPSG-based parsing's performance advantages. 
Specifically, we define a novel learnable target for the head parameter and add a separate scorer to the model to predict the score of a position being a head. 
By inputting this score to the modified HPSG joint decoding algorithm, we expedite the expensive decoding to $O(n^3)$.
We conducted extensive experiments on the PTB and CTB datasets and the SPMRL14 and UD benchmarks to verify the effectiveness of our proposed method. 
The experimental results on monolingual parsing show that we achieve comparable or even better performance while having better decoding efficiency. 
In the multilingual benchmark, the proposed model achieves the new state-of-the-art.
In addition, because HPSG-based parsing requires parallel constituent and dependency parallel syntactic treebanks during training, we discussed the impact of different conversions for dependency trees, as dependency treebanks can be easily converted from constituent treebanks using head rules.
Conversely, the dependency to constituency treebank conversions usually require machine learning approaches, so we proposed a simple conversion algorithm based on HFP, which supports both projective and non-projective dependency trees, and explored this as an option for when only a dependency treebank is available.

\begin{figure*}[t]
    \centering
    \includegraphics[width=0.8\textwidth]{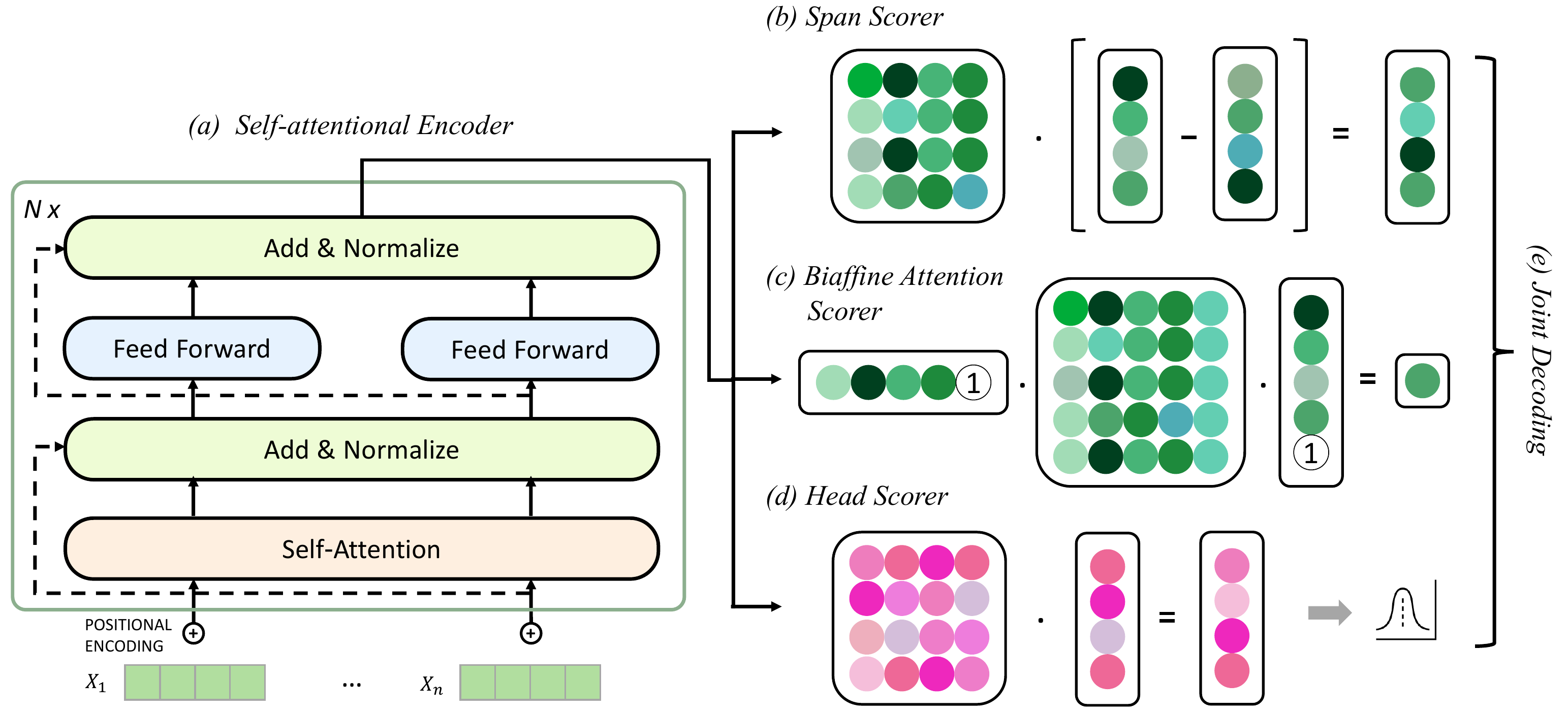}
    \caption{The system architecture of our proposed HPSG-based parsing model.}
    \label{fig:arch}
\end{figure*}

\section{Background}\label{sec:background}

The neural HPSG-based parser \cite{zhou-zhao-2019-head} is an approach for jointly parsing constituent and dependency trees using a simplified HPSG tree structure. 
In the training phase of this parser, two training objectives are optimized: Chart Structure Prediction (CSP) \cite{stern-etal-2017-minimal} and Dependency Head Prediction (DHP) \cite{dozat2016deep}. These two objectives jointly train the parser's encoder parameters.
In this process, CSP optimizes the model for chart-based constituent parsing, and DHP optimizes the model for graph-based dependency parsing.
In the inference phase, a novel HPSG-based decoding algorithm is used to jointly decode the constituency and dependency parse trees simultaneously.

\paragraph{Formalization} 
Given a sequence of tokens $X = (x_1, x_2, \ldots, x_n)$, a constituency tree can be defined as a collection of labeled spans over these tokens in chart-based parsing: $\mathcal{C} = \{(i, j, \ell^c), 1 \leq i \leq j \leq n, \ell^c \in \mathcal{L}^c\}$, where $\mathcal{C}$ represents a constituent located between fencepost positions $i$ and $j$ in a sentence and with label $\ell^c$.
A dependency parse tree for the sentence can be defined using dependents $\mathcal{D} = \{(h, m, \ell^d), 0 \leq h \leq n, 1 \leq m \leq n, \ell^d \in \mathcal{L}^d\}$, where $(h, m, \ell^d)$ consists of a governor (head word, $x_h$), a dependent (modifier word, $x_c$), and the type of the relation $\ell^d$, which is in label set $\mathcal{L}^d$. In dependency parsing, a pseudo token $x_0$ is used to represent the root node of the parse tree. %(refer to Appendix \ref{sec:background} for more details of HPSG-based parsing)

The score of a constituent tree $T_c$, denoting the tree's likelihood, is defined as the sum of its constituent label and span scores:
\begin{equation*}
    \footnotesize
	\begin{aligned}
		s(T_c) &= \sum_{(i,j,\ell^c) \in T_c} s(i, j, \ell^c) \\
		& = \sum_{(i,j,\ell^c) \in T_c} [s_{span}(i, j) + s_{label}((i, j), \ell^c)]
	\end{aligned}
	\label{eq:s_tc}
\end{equation*}
where $s(i, j, \ell^c)$ is real-valued score for a constituent. The training goal of constituency parsing is to find the highest-scoring constituent tree:
\begin{equation*}
    \footnotesize
	\begin{aligned}
		\hat{T}_c = \arg\max_{T_c}[s(T_c)],
	\end{aligned}
\end{equation*}
with golden parse tree $T_c^*$. The model trains to satisfy the following margin constraint:
\begin{equation*}
    \footnotesize
	\begin{aligned}
		s(T_c^*) - s(\hat{T}_c) \leq \Delta(T_c^*, \hat{T}_c),
	\end{aligned}
\end{equation*}
in which $\Delta(\cdot)$ is the constituent tree Hamming loss, which measures the similarity between predicted and reference trees. To encourage this Hamming loss constraint and reduce the magnitude of the largest margin violation, the hinge loss is minimized:
\begin{equation*}
    \footnotesize
	\begin{aligned}
	    \mathcal{J}_c(\theta) = \max \big(0, \max_{T^*_c \neq \hat{T}_c}[s(\hat{T}_c) + \Delta(T_c^*, \hat{T}_c)] - s(T^*_c) \big),
	\end{aligned}
\end{equation*}
%In addition, the $s(i, j, \ell^c)$ in Eq. (\ref{eq:s_tc}) is replaced with $s(i, j, \ell^c) + \mathbf{1}[\ell^c \neq l^{c*}_{i,j}]$, where $l^{c*}_{i,j}$ is the label of span $(i, j)$ in the gold tree $T^*_c$ for enlarging the margins between the gold tree and predictions that contain more mistakes and further offering a more robust and generalized training.

Notably, in chart-based constituent parsing, the empty label $\varnothing$ is included in the set of constituent labels to binarize each $n$-ary subtree in the training data, and the unary chains in the data are collapsed into single entries. 

In dependency parsing, the score of dependency tree $T_d$ is also composed of two parts: the scores of the head-modifier arcs and the scores of the arcs' labels. Formally, it is defined as:
\begin{equation*}
    \footnotesize
	\begin{aligned}
		s(T_d) &= \sum_{(h,m,\ell^d) \in T_d} s(h, m, \ell^d) \\
		& = \sum_{(h,m,\ell^d) \in T_d} [s_{arc}(h, m) + s_{label}((h, m), \ell^d)].
	\end{aligned}
\end{equation*}
Although the scoring equations for dependency and constituent parsing are consistent, since the number of modifiers ($m = [1, 2, ..., n]$) in the dependency tree representation is known and fixed, graph-based dependency parsing simplifies to the problem of finding the connections between these modifiers. 
Therefore, a negative log likelihood of the gold tree $T_d^*$ is minimized as the training loss:
\begin{equation*}
    \footnotesize
	\begin{aligned}
		\mathcal{J}_d(\theta) = \sum_{m=1}^{n} [s_{arc}(h^*, m) + s_{label}((h^*, m), l^{d*})],
	\end{aligned}
\end{equation*}
where $h^*$ is the head of modifier $m$ and $l^{d*}$ is the label of pair $(h^*, m)$ in the gold tree $T_d^*$.

\section{Our Parser}

In this work, we adopt the state-of-the-art neural HPSG-based parser \cite{zhou-zhao-2019-head} as our basic parsing framework.
Figure \ref{fig:arch} shows the overall framework of our proposed parser. 
The parser is composed of five parts: the self-attentional encoder, the biaffine attention scorer, the chart scorer, the head scorer, and the joint decoding module. 
We adopt a same self-attentional encoder same as in \cite{zhou-zhao-2019-head}, the input tokens are encoded into contextualized representations $y$.
%(refer to Appendix \ref{sec:encoder} for a detail explanation)
%kitaev-klein-2018-constituency,

\subsection{Self-attentional Encoder}\label{sec:encoder}

Self-attention, especially multi-head self-attention \cite{vaswani2017attention}, has show great success in a variety of tasks, including constituent parsing \cite{kitaev-klein-2018-constituency} and dependency parsing \cite{li2019self}. 
In self-attention, an input layer maps each input token $x_i$ in position $i$ and its corresponding tag $t_i$ into a dense vector representation $e_i$:
\begin{equation}
    \footnotesize
	\begin{aligned}
	e_i = (\mathbf{emb}^{word}_{x_i} \oplus \mathbf{emb}^{tag}_{t_i}) \oplus \mathbf{emb}_{i}^{pe},
    \end{aligned}
\end{equation}
where $\mathbf{emb}^{word}$, $\mathbf{emb}^{tag}$, and $\mathbf{emb}^{pe}$ are learnable word embeddings, tag embeddings, and position embeddings, respectively. All share the same dimensionality, $d_{model}$.

The self-attentional encoder then encodes this dense vector representation into contextualized representations $y_i$ using $N$ identical stacked self-attention layers with $M$ separate heads each:
\begin{equation}
    \footnotesize
	\begin{aligned}
    \mathbf{MultiHead}(H) &= \sum_{k=1}^M \mathbf{SelfAttn}^k(H),\\
    \mathbf{SelfAttn}(H) &= \big[\mathbf{softmax}(\frac{QK^T}{\sqrt{d_k}})\big]V,
	\end{aligned}
\end{equation}
where $d_k = d_{model} / M$ is the dimension of each head. $H$ is the hidden states, $Q = W_Q H$, $K = W_K H$, and $V = W_V H$.
After multi-head attention, residual connections and Layer Normalization \cite{ba2016layer} are adopted for better training.

A feed-forward layer (FFN) with a nonlinear activation function, the Rectified Linear Unit (ReLU), is adopted to transform the outputs from multi-head attention to reduce load on the model, and then a residual connection and layer normalization are again used.

\subsection{Span Scorer}
%As presented in the Appendix \ref{sec:background}, 
In HPSG-based parser, the training of the constituent parse tree requires scoring spans.
We adopt a span scorer based on the contextualized output of the self-attentional encoder following the pratice in \cite{gaddy-etal-2018-whats,kitaev-klein-2018-constituency}.
In the span scorer, the score is calculated based on the span features which are derived using span boundaries. 
Specifically, for span $(i, j)$, the feature $f^s_{i,j}$  is obtained using contextualized vectors in both forward and backward directions of the boundaries:
\begin{equation*}
    \footnotesize
	\begin{aligned}
	f^s_{i,j} = (\overrightarrow{y}_j - \overrightarrow{y}_{i-1}) \odot (\overleftarrow{y}_{j+1} - \overleftarrow{y}_i),
	\end{aligned}
\end{equation*}
where $\odot$ represents the concatenation operation, $y$ are contextualized outputs from the self-attentional encoder, and $\overrightarrow{y}$ and $\overleftarrow{y}$ the first half and second half of $y$, respectively. 

Using the obtained span features, a single layer FFN with ReLU activation and Layer Normalization generates the span scores vector:
\begin{equation*}
    \footnotesize
	\begin{aligned}
	s_{span}(i, j) = \mathbf{FFN}(f^s_{i,j}),
	\end{aligned}
\end{equation*}
where the scores vector $s_{span}(i, j)$ has length $|\mathcal{L}_c|$. The score predicted for label $\ell^c$ is then
{\footnotesize $s_{label}((i, j), \ell^c) = [s_{span}(i, j)]_{\ell^c}$}.

\subsection{Biaffine Attention Scorer}

Since the phrase and dependency structure are jointly represented using a recursive \textit{joint span} structure in the HPSG-based parser, in addition to the span scorer, an additional dependency structure scorer is required. 
We adopt an biaffine attention scorer to compute the scores of all dependencies \cite{dozat2016deep}.

First, to obtain role-specific representations and strip the irrelevant information, MLPs are applied to the contextualized representation in this scorer:
\begin{equation*}
    \footnotesize
	\begin{aligned}
	 y^H_i = \mathbf{MLP}^H(y_i), \quad y^M_i = \mathbf{MLP}^M(y_i).
	\end{aligned}
\end{equation*}

The biaffine attention computes a head-modifier pair's score as:
\begin{equation*}
    \footnotesize
	\begin{aligned}
		s_{arc}(h, m) = \Big[
		\begin{array}{c}
		     y^M_m  \\
		     1
		\end{array}
		\Big] \cdot W_{biaf}  \cdot y^H_h,
	\end{aligned}
\end{equation*}
and all candidate head-modifier pairs are scored simultaneously using a matrix form of this equation.

\subsection{HPSG-based Joint Decoding}

For only parsing constituent trees, a dynamic program algorithm is adopted in decoding. 
$s_{best}$ is defined as the overall best score for a labeled span.
The best score $s_{best}$ for a minimum span $(i,i+1)$ is only a single maximization over the label candidates since the span scores do not need to be considered.
\begin{equation*}
    \footnotesize
	\begin{aligned}
	s_{best}(i, i+1) = \max_{\ell^c}\big(s_{label}((i, i+1), \ell^c)\big)
    \end{aligned}
\end{equation*}
And the $s_{best}$ for general span $(i, j)$ is recursively defined on the sub-spans:
\begin{equation*}
    \footnotesize
	\begin{aligned}
	s_{best}(i, j) = &\max_{\ell^c, k}\big(s_{label}((i, j), \ell^c) \\
	&+ s_{span}(i, k) + s_{span}(k, j) \\
	&+ s_{best}(i, k) + s_{best}(k, j)\big)
    \end{aligned}
\end{equation*}
in which the split score for span $(i, j)$ with split position $k$ is {\footnotesize $s_{split}(i, k, j) = s_{span}(i, k) + s_{span}(k, j) + s_{best}(i, k) + s_{best}(k, j)$}.

For HPSG-based parsing, a true joint decoding process is adopted instead of a joint learning and separate decoding paradigm. Constituent and dependency structures are considered and generated at the same time in this joint decoding.
Based on the joint span structure in HPSG trees, the $s_{best}$ for general span $(i, j)$ becomes:
\begin{equation*}
    \footnotesize
	\begin{aligned}
	s_{best}(i, j) = &\max_{\ell^c, k}\big(s_{label}((i, j), \ell^c) \\
	&+ s_{span}(i, k) + s_{span}(k, j) \\
	&+ s^{hpsg}_{best}(i, k, j) \big)
    \end{aligned}
\end{equation*}
in which $s^{hpsg}_{best}(i, k, j)$ indicates the best HPSG structure score of span $(i, j)$ with split position $k$ and is defined as:
\begin{equation}
    \footnotesize
	\begin{aligned}
	s^{hpsg}_{best}(i, k, j) = \max_{\ell^d, k}\Big(~~~~~~~~~~~~~~~~~~~~~~~~~~~~~~~~~~~~~~~~~~~~~~~~~~~~~~~~~~~~& \\
	\max\big[
	\max_{i \leq m \leq k < h \leq j}(s_{best}(i, k) + s^*_{best}(k, j) + s_{arc}(h, m)), \\
	\max_{i \leq h \leq k < m \leq j}(s^*_{best}(i, k) + s_{best}(k, j) + s_{arc}(h, m))\big]\\
	+ s_{label}((h, m), \ell^d) \Big)~~~~~~~~~~~~~~~~~~~~~~~~~~~~~~~&
    \end{aligned}
    \label{eq:hpsg}
\end{equation}
where $s_{arc}(h, m)$ is dependency score between the head word\footnote{The head here is not the same as the head in the dependency parsing. The head in HPSG refers to the head word in a phrase/span. According to the HFP, the head word serves as a dependency relation's modifier for words outside the span but a dependency head for words inside the span.} $h$ and $m$ of sub-spans $(i, k)$ and $(k, j)$, $s^*_{best}(\cdot)$ is the version of $s_{best}(\cdot)$ without empty labels, that is {\footnotesize $s^*_{best}(\cdot) = \max\limits_{\ell^c \neq \varnothing, k} (\cdot)$}.

When only parsing constituent trees, a recursive CKY algorithm \cite{stern-etal-2017-minimal} is used to find the parse tree with the highest score for a given sentence. 
In short, it takes $O(n^2)$ time to look for the start and end for a span, and $O(n)$ time to look for the split position, so the total running time of this proposed algorithm is $O(n^3)$, the same as classical chart parsing.
In the HPSG-based parsing, when scoring the HPSG structures, identifying the head words of pairs of sub-spans is crucial. This is done using dependency scores and following the HFP. 
The extra time overhead of this part is $O(n^2)$, so the time complexity of HPSG-based joint decoding reaches an expensive $O(n^5)$. 
While costly, this joint decoding process bypasses separately using CKY and MST algorithms for decoding. 
Additionally, it concurrently generates the output for these two syntactic parse tree types while outperforming other decoding processes; however, the cost still motivates us to pursue further improvements. 

\subsection{Head Scorer}

We revisit the role of the head word in order to reduce the cost of searching for the head word in the hopes of reducing the time complexity of the HPSG-based joint decoding algorithm to be closer to that of the original CKY algorithm. 
In the original implementation of the HPSG-based joint decoding algorithm, the head word is used as a structure constraint; i.e., the head words of two sub-spans that are split should obey the HFP. An exhaustive search based on dependency score is then adopted to find the head word and compose part of the overall score.
Rather than searching through dependency scores, we propose a novel head scorer that outputs the score of a head word directly. 

Since the head score $s_{head}$ is not a simple binary score to indicate whether the word is a head or not, it needs to satisfy certain properties. We pick $\hbar$ to signify the position of a head word to better differentiate the head term $h$ in dependency parsing, and $\hbar_{(i, j)}$ means the head of span $(i, j)$.

\noindent\textbf{\textit{Prop 1:}} The head score of the head word is higher than the head score of any non-head word in the span.
$$s_{head}(\hbar) \geq s_{head}(\psi), \forall i \leq \psi \leq j$$
\noindent\textbf{\textit{Prop 2:}} The score for a head word of multiple spans remains consistent across spans.
$$s_{head}(\hbar_{(i_1, j_1)}) = s_{head}(\hbar_{(i_2, j_2)}), \hbar_{(i_1, j_1)} = \hbar_{(i_2, j_2)}$$
\noindent\textbf{\textit{Prop 3:}} A span has a head score greater than or equal to the head scores of all of its sub-spans.
$$s_{head}(\hbar_{(i_1, j_1)}) \geq s_{head}(\hbar_{(i_2, j_2)}), i_1 \leq i_2 \wedge j_1 \geq j_2$$

\begin{table*}[t]
	\centering
	\small
	\setlength{\tabcolsep}{3pt}
	\caption{Constituent and dependency parsing on PTB and CTB 5.1 test datasets. Results of BenePar + BERT are reported in \cite{kitaev-etal-2019-multilingual}. HAcc. is abbreviation for head accuracy. Speed refers to average sentences per second on the test set. $^*$ indicates the results of the dependency parsing on the CTB5.1 constituent parsing split.}\label{tab:ptb_ctb} 
	%\begin{sc}
	\begin{tabular}{lcc>{\columncolor[rgb]{0.95,0.95,0.95}}ccc>{\columncolor[rgb]{0.95,0.95,0.95}}ccc>{\columncolor[rgb]{0.95,0.95,0.95}}ccc>{\columncolor[rgb]{0.95,0.95,0.95}}c}
		\toprule
		\multirow{2}{*}{\bf Model} & \multicolumn{7}{c}{PTB} & \multicolumn{5}{c}{CTB5.1} \\
		\cmidrule(lr){2-8} \cmidrule(lr){9-13} & HAcc. &UAS &LAS & LP & LR & LF$_1$ & Speed  &UAS &LAS & LP & LR & LF$_1$ \\
		\midrule
		Kuncoro et al. \cite{kuncoro-etal-2017-recurrent} & $-$ & 95.80 & 94.60 & $-$ & $-$ & 93.60 & $-$  & $-$ & $-$ & $-$ & $-$ & $-$  \\
		Dozat and Manning \cite{dozat2016deep} & $-$  &95.74 &94.08 & $-$ & $-$ & $-$ & $-$  &89.30 &88.23 & $-$ & $-$ \\
		BenePar \cite{kitaev-klein-2018-constituency} & $-$  & $-$ & $-$ & 93.20 & 93.90 & 93.55 & $-$  & $-$ & $-$ & $-$ & $-$ & $-$ \\
		\quad + BERT & $-$  & $-$ & $-$ & 95.46 & 95.73 & 95.59 & 241.6 & $-$  & $-$ & 91.55 & 91.96 & 91.75 \\
		HPSG-Par \cite{zhou-zhao-2019-head}  & $-$  & 96.09 & 94.68 & 93.64 &  93.92 &  93.78 & $-$ &  91.21 &89.15 & 89.09 & 89.70 & 89.40\\
		\quad + BERT/RoBERTa  & $-$  & 97.00 & 95.43 & 95.70 & 95.98 & 95.84  & $-$ & $-$ & $-$ & 92.50 & 92.61 & 92.55 \\
		\quad + XLNet & $-$  & 97.20 & 95.72 & 96.21 & 96.46 & 96.33 & 158.7 & $-$ & $-$ & $-$ & $-$ & $-$  \\
		\hdashline
		\makecell[l]{SAPar \cite{tian-etal-2020-improving-constituency} \\(+ BERT)} & $-$ & $-$ & $-$ & 96.09 & 95.62 & 95.86 & $-$ & $-$ & $-$ & 92.83 & 92.50 & \bf 92.66 \\
		(+ XLNet/ZEN) & $-$  & $-$ & $-$ & 96.61 & 96.19 & \bf 96.40 & $-$ &  $-$ & $-$ & 92.61 & 92.42 & 92.52 \\
		\hdashline
		\makecell[l]{KMPar \cite{mrini-etal-2020-rethinking}\\(+ XLNet/BERT)} & $-$  & \bf 97.42 & \bf 96.26 & 96.24 & 96.53 & 96.38 & $-$ & 94.56$^*$ & 89.28$^*$ & 91.85 & 93.45 & 92.64 \\
		\midrule
        \bf H3n-Par & 64.89 & 95.90 & 94.57 & 93.40 & 93.63 & 93.51 & 234.4 & 90.98 & 88.85 & 89.10 &	88.92 & 89.00\\
        \quad + BERT/RoBERTa & 68.50 & 96.97 & 95.27 & 95.69 & 95.42 & 95.56 & 170.5 & 91.15 & 89.04 & 92.35 & 92.18 & 92.26 \\
        \quad + XLNet & 68.95 & 97.31 & 95.76 & 96.26 & 96.15 & 96.20 & 148.5  & \makecell[c]{\bf 91.86 \\ 93.89$^*$} & \makecell[c]{\bf 89.71 \\ 92.34$^*$} & 92.66 & 92.52 & 92.59 \\
		\bottomrule
	\end{tabular}
	%\end{sc}
	
\end{table*}

As long as these three properties are met, the deterministic decoding phase of the HPSG structure can be guaranteed. Amongst these, Prop. 1 gives the heuristic for finding a head word within a span, Prop. 2 illustrates the phenomenon of head sharing in the HPSG structure of the stacked spans, and the combination of Prop. 2 and Prop. 3 ensures that decoding does not have to perform an exhaustive search over dependency arcs if a span is split into sub-spans.

In researching head words, we found that the level of a span in a HPSG tree provides very significant information for head score calculation.
We define the level of node (span) $(i,j)$ with head $\hbar$ in the HPSG tree as $sl(i,j)$. This is equal to 1 plus the number of arcs experienced in traversing from the root to that node along the tree. Then the gold head score $s^{gold}_{head}$ can be written as:
\begin{equation*}
    \footnotesize
	\begin{aligned}
	s^{gold}_{head}(\psi) = \Big\{\begin{array}{rl}
	     \frac{1}{\min_{(i, j)}sl(i,j)},& \psi \in [\hbar] \\
	     0,&  \text{otherwise}
	\end{array}
    \end{aligned}
\end{equation*}

Our preliminary experiments that took head scoring as a regression task with $s^{gold}_{head}(\psi)$ as the target performed worse than taking $sl(i,j)$ as the target in a classification task. 
Therefore, we define the level of a word belonging to no span as +INF, and then limit the maximum level to fix the number of possible classification targets.
We adopt a feed-forward network layer as the head scorer. The final head score is given:
\begin{equation*}
    \footnotesize
	\begin{aligned}
	s_{head}(\psi) = \frac{1}{\arg\max(\mathbf{FFN}(y_\psi))}.
	\end{aligned}
\end{equation*}
%We present the detailed implementation in Appendix \ref{sec:decoding_algo}.

\begin{algorithm}[t!]
    \small
	\caption{H3n Joint Decoding algorithm}	\label{alg:h3n}
	\begin{algorithmic}
		\REQUIRE sentence length $n$, constituent span $s^\clubsuit(i,j,\ell)$, dependency arc $s^\spadesuit(h,m)$, and head score $s^\heartsuit(i)$, $1\leq i\leq j \leq n, \forall h,m,\ell$
		\ENSURE maximum value $S_H(T)$ of tree $T$ \\
		\STATE \textbf{Initialization:} 
		\STATE $s_{c}[i][j] = s_{i}[i][j] = 0, hs[i][j] = 0, \forall i,j $ 
		\FOR{$len=1$ to $n$}
		\FOR{$i=1$ to $n-len + 1$}
		\STATE $j = i + len - 1$
		\IF {$len=1$}
		\STATE $\begin{aligned}
			s_{c}[i][j][i] &= s_{i}[i][j][i] = \max_{\ell} s^\clubsuit(i,j,\ell) \\
			hs[i][i] &= i
		\end{aligned}$
		\ELSE
		\FOR{$k=i$ to $j$}
		\STATE 
		$\begin{aligned}
			head_l &= s^\heartsuit(hs[i][k]) \\
			head_r &= s^\heartsuit(hs[k+1][j]) \\
		\end{aligned}$
		\IF {$head_l > head_r$}
		\STATE 
		$\begin{aligned}
			split &=  s_{c}[i][k] + s_{i}[k+1][j] + \\
			&s^\spadesuit(hs[i][k],hs[k+1][j]) \\
			hs[i][j] &=  hs[i][k]
		\end{aligned}$
		\ELSE
		\STATE  
		$\begin{aligned}
			split &=  s_{i}[i][k] + s_{c}[k+1][j] + \\
			&s^\spadesuit(hs[k+1][j],hs[i][k]) \\
			hs[i][j] &=  hs[k+1][j]
		\end{aligned}$
		\ENDIF
		\STATE $\begin{aligned}
			s_{c}[i][j] = & \max \ \{\ s_{c}[i][j], split + \\ 
			&\max_{\ell \neq \varnothing} s^\clubsuit(i,j,\ell) \ \} 
		\end{aligned}$
		\STATE $\begin{aligned}
			s_{i}[i][j] = &\max \ \{\ s_{i}[i][j], split + \\ 
			& \max_{\ell} s^\clubsuit(i,j,\ell) \ \}
		\end{aligned}$
		\ENDFOR
		\ENDIF
		\ENDFOR
		\ENDFOR
		\STATE $\begin{aligned}
			S_H(T) =   s_{c}[1][n] + d(hs[1][n],root)
		\end{aligned}$
	\end{algorithmic}
\end{algorithm}

\subsection{Our Decoding Algorithm}

Our decoding algorithm, H3n (HPSG with 3 power of n time complexity), is based on the original HPSG-based joint decoding, The only difference is in the computation of $s^{hpsg}_{best}(i, k, j)$. In our decoding algorithm, $s^{hpsg}_{best}(i, k, j)$ for span $(i, j)$ with split position $k$ is computed as:
\begin{equation}
    \footnotesize
	\begin{aligned}
	s^{hpsg}_{best}(i, k, j) = \max_{\ell^d, k}\Big( s_{label}((h, m), \ell^d)  \quad\quad\quad\quad\quad\quad\quad&\\
	+ \Bigg\{\begin{array}{cc}
	     s_{best}(i, k) + s^*_{best}(k, j) + s_{arc}(\hbar_{(k, j)}, \hbar_{(i, k)}),  & \\
	     \quad\quad\quad\quad\quad\quad s_{head}(\hbar_{(k, j))} > s_{head}(\hbar_{(i, k))} &  \\
	     & \\
	     s^*_{best}(i, k) + s_{best}(k, j) + s_{arc}(\hbar_{(i, k)}, \hbar_{(k, j)}), & \\
	     \quad\quad\quad \text{otherwise} &
	\end{array} \Bigg)&
    \end{aligned}
    \label{eq:hpsg1}
\end{equation}

Comparing Eq.(\ref{eq:hpsg}) and (\ref{eq:hpsg1}), due to the introduction of the head score, $s_{head}$, the cost of the $O(n^2)$ head word search and comparison process is reduced. 
By using head scores, the head word search can be combined with the split point scanning process, which reduces the time complexity of our H3n decoding algorithm to $O(n^3)$, and is equivalent to that of the CKY algorithm.

%\section{Our Proposed H3n Joint Decoding Algorithm}\label{sec:decoding_algo}

In terms of implementation, for efficient decoding, we propose a CKY-style algorithm based on \cite{zhou-zhao-2019-head} to explicitly find the globally highest span and dependency score of the simplified HPSG tree, as shown in Algorithm \ref{alg:h3n}. 
In order to binarize constituent parse trees with heads, we introduce the complete span $s_c$, the incomplete span $s_i$, and head score $hs$, which is similar to the Eisner algorithm \cite{eisner-1996-efficient}. After finding the best score, we backtrack the chart with the points and dependency arcs to construct both a constituent and a dependency tree.

\subsection{Pseudo-constituent Tree Synthesis}\label{sec:restore_const}

\begin{algorithm}[t!]
  \small
  \caption{Convert DepTree to ConstTree}\label{alg:deptree2consttree}
  \begin{algorithmic}
  \REQUIRE dependency tree $T_d$
  \ENSURE constituent tree $T_c$
  \STATE $flag = 0, childlist = empty $ 
  \FOR{$i$ in each son of the $root$ of $T_d$}
    \IF {$i$ is behind of $root$ and $flag == 0$}
    \STATE make constituent tree leaf $l$ of $root$ and put $l$ into $childlist$
    \STATE $flag = 1$
    \ENDIF
    \STATE apply recursive call to convert DepTree to ConstTree for son $i$, and put constituent tree $T_c^i$ of son $i$ into $childlist$
  \ENDFOR
  \IF {$flag == 0$}
  \STATE make constituent tree leaf $l$ of $root$, then put $l$ into $childlist$
  \ENDIF
  \STATE make constituent tree $T_c$ of $childlist$ 
  \FOR{$j$ in each internal node of the $T_c$ from bottom to top}
      \WHILE{$j.children[-1].id - j.children[0].id \neq j.children.length$}
          \FOR{$jc$ in $j.children$}
          \STATE $tmp = j.parent$
              \WHILE{$tmp.parent \neq none$}
                  \IF{$jc.id < tmp.left$ or $jc.id > tmp.right$}
                  \STATE move $jc$ to be a child of $tmp$
                  \STATE break
                  \ENDIF
              \ENDWHILE
          \ENDFOR
      \ENDWHILE
  \ENDFOR
  \STATE return constituent tree $T_c$
  \end{algorithmic}
\end{algorithm}

Since the HPSG structure requires parallel constituent and dependency structures for training, in order to support the training of the HPSG parsing model when only the dependency structure is available, we propose an algorithm for generating constituent trees from dependency trees. 
Different from some previous algorithms, our algorithm does not require a supervised machine learning approach, and the pseudo-constituent trees are also annotated with syntactic labels.
Specifically, we use the relationship between the head word in the phrase and its governor in the dependency tree as the constituent label of the phrase. The conversion process is shown in Algorithm \ref{alg:deptree2consttree}.

The first step of this algorithm is to use all the children of the dependency subtree to form a phrase according to the HFP. 
This idea is simple and meets the requirements of HPSG structure. 
In order to be able to adapt to a variety of dependency tree annotation standards such as Stanford Dependency or UD, we do not adopt a conversion method based on dependency relation and grammar rules. 
For the derived the phrase structure, the relationship between the head words of these phrases and their dependency parent is regarded as the syntactic category of the phrase.

\begin{figure}[t!]
    \centering
    \includegraphics[width=0.5\textwidth]{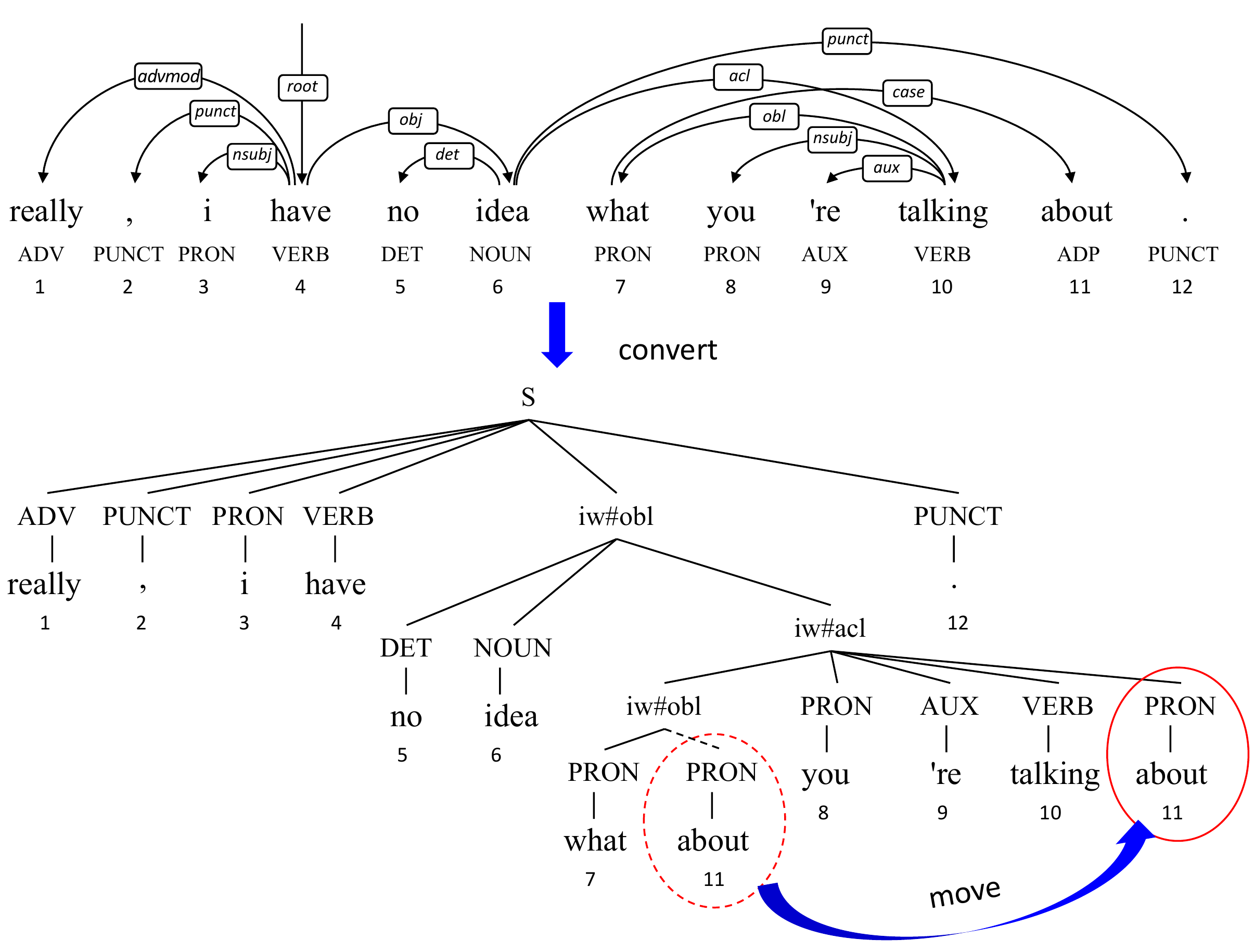}
    \caption{An example showing the conversion process.}
    \label{fig:tree_conversion}
\end{figure}

The existing dependency to constituent tree conversion methods rarely consider the influence of non-projective dependency structure. 
Due to the existence of non-projective dependent arcs, the relative order of many leaf nodes in the obtained phrase structure changes. 
Therefore, when finding these phrases, we try to gradually move the child nodes of these phrases to their parent phrases so that leaf nodes conform to their original order in the end.
Figure \ref{fig:tree_conversion} gives an example of converting a non-projection dependency tree. If no subsequent correction is made, the constituent tree obtained by relying solely on the HFP is incorrect.

\section{Experiments}

\subsection{Setup}

Our experiments include three categories: classical parsing evaluation benchmarks (the Penn Treebank 3 (PTB) \cite{marcus-etal-1993-building} and the Chinese Penn Treebank (CTB) 5.1 \cite{xue2005penn}), a multilingual constituent parsing benchmark (the Statistical Parsing of Morphologically Rich Languages (SPMRL) 2014 shared task \cite{seddah-etal-2014-introducing}), and a multilingual dependency benchmark (Universal Dependencies (UD) v2.6)\footnote{http://hdl.handle.net/11234/1-3226}.
For PTB, we follow the standard splits \cite{zhang-clark-2008-tale,liu-zhang-2017-order} for both constituency and dependency parsing with sections \textit{2–21, 22, 23} as the training, development, and test sets, respectively.
For CTB, since the data splitting widely used in constituent parsing (\textit{train: 1-270, 440-1151, dev: 301-325, test: 271-300}) \cite{petrov-klein-2007-improved,liu-zhang-2017-order} differs from that used in dependency parsing (\textit{train: 1-815, 1001-1136, dev: 886-931, 1148-1151, test: 816-885, 1137-1147}) \cite{zhang-clark-2008-tale}, we report results for constituent and dependency parsing separately on their respective splits.
In the experiments for PTB and CTB, we use Stanford Parser \cite{de-marneffe-etal-2006-generating} 3.3 to convert constituent trees to dependency trees and were consistent with \cite{zhou-zhao-2019-head} unless otherwise stated.
Predicted part-of-speech tags in PTB and CTB are provided by the Stanford tagger \cite{toutanova-etal-2003-feature}.
In SPMRL, the Universal Dependencies 1.0 head rules\footnote{https://nlp.stanford.edu/software/stanford-dependencies.shtml} are used to convert the constituent trees into dependency trees to generate a simplified HPSG structure for training. %
In UD, we introduce an algorithm based on the HFP to help restore both projective and non-projective dependency trees to pseudo-constituent trees.
% (refer to Appendix \ref{sec:restore_const} for a detailed explanation)

\begin{table}[]
    \centering
    \small
    \caption{Decoding time with the same 1,000 Wikipedia sentences in varying length conditions. Running time comes from average of repeatedly decoding 5 times. The machine was Intel Xeon Gold 6150, NVIDIA TESLA V100 GPU, with batch size 50.}
    \label{tab:speed}
    %\resizebox{1.0\linewidth}{!}{
    \begin{tabular}{ c | l | c | r r r }
     \toprule
     Length & Algo. & Comp. & Time & Speed & Speedup  \\ 
     \midrule
      \multirow{4}{*}{\makecell[c]{0-100 \\(avg: 26.6)}} & CKY  & $O(n^3)$ & 12.5s & 80.0 & $-$ \\
      &MST & $O(n^3)$ & 14.2s & 70.4 & $-$ \\
      &HPSG & $O(n^5)$ & 23.7s  & 42.2 & $-$ \\
      &\bf H3n & $O(n^3)$ & 16.8s & 59.5 & 1.41x \\
    \midrule
     \multirow{4}{*}{\makecell[c]{100-200 \\(avg: 127.0)}} & CKY & $O(n^3)$ & 60.5s  & 16.5 &$-$ \\
     &MST & $O(n^3)$ & 62.3s & 16.1 & $-$ \\
     &HPSG & $O(n^5)$ & 2014.2s & 0.5  & $-$ \\
     &\bf H3n & $O(n^3)$ & 113.5s & 8.8 & 17.7x \\
     \midrule
     \multirow{4}{*}{\makecell[c]{200-300 \\(avg: 232.6)}} & CKY  & $O(n^3)$ & 134.6s & 7.4 & $-$ \\
     &MST & $O(n^3)$ & 136.5s & 7.3 & $-$ \\
     &HPSG & $O(n^5)$ & 179882.3s & 0.006 & $-$ \\
     &\bf H3n & $O(n^3)$ & 281.5s & 3.6 & 639.0x \\
     \bottomrule
    \end{tabular}
    %}
    
\end{table}

\begin{table*}[t]
\centering
\setlength{\tabcolsep}{3pt}
\caption{Results on the test datasets of the SPMRL14 shared task. F$_1$ scores are from its official \texttt{evalb} script. $^\dag$ indicates dependency trees from the original SPMRL14, while $^\ddag$ indicates the dependency trees using the UD 1.0 head rules. The result of Polish with UD conversion is unavailable due to its non-standard constituent labels. The scores in parentheses are the averaged without Polish for comparison.}
\label{table:spmrl}
%\resizebox{1.0\linewidth}{!}{
\begin{tabular}{@{}lcccccccccccr}
\toprule
  & &Arabic &Basque &French &German &Hebrew &Hungarian &Korean &Polish &Swedish & Average \\
  \midrule
  Bjorkelund et al. \cite{bjorkelund-etal-2014-introducing} & F$_1$ & 81.32 &88.24&82.53&81.66&89.80&91.72&83.81&90.50&85.50&$~~~~~~~~~~~~$ 86.12  \\
  Coavoux and Crabbe \cite{coavoux-crabbe-2017-multilingual} & F$_1$ & 82.92 &88.81&82.49&85.34&89.87&92.34&86.04&93.64&84.00&$~~~~~~~~~~~~$  87.27 \\
  Kitaev et al. \cite{kitaev-etal-2019-multilingual} & F$_1$ &85.61&89.71&84.06&87.69&90.35&92.69&86.59&93.69&84.45&$~~~~~~~~~~~~$  88.32 \\ 
  \quad+ mBERT & F$_1$ & \bf 87.97 & 91.63  & 87.42  & 90.20  & 92.99  & 94.90  &88.80 & 96.36 &88.86 & $~~~~~~~~~~~~$  91.01 \\  
  \midrule
  \multirow{3}{*}{\bf HPSG-Par + mBERT$^\dag$} & F$_1$ & 87.90 & 92.14 & 87.27 & 90.36 & 93.17 & \bf  95.44 & \bf 89.53 & \bf 96.55 & 89.52 & $~~~~~~~~~~~~$ \bf 91.32  \\
  & UAS & 91.84 & 91.62 & 91.96 & 91.94 & 91.00 & 92.35 & 92.58 & 94.85 & 91.83 & $~~~~~~~~~~~~$ 92.22 \\
  & LAS & 90.24 & 87.46 & 88.53 & 90.14 & 85.21 & 88.89 & 90.73 & 91.37 & 87.69 & $~~~~~~~~~~~~$  88.92 \\
  \hdashline
  \multirow{3}{*}{\bf H3n-Par + mBERT$^\dag$} & F$_1$ & 87.63 & 91.92 & 87.52 & 90.13 & \bf 93.24 & 95.36 & 89.21 & 96.52 & 89.41 &  (90.55) 91.22 \\
  & UAS & 91.46 & 91.21 & 92.06 & 90.43 & 91.06 & 92.09 & 92.34 & 94.85 & 91.80 & (91.56) 91.92 \\
  & LAS & 89.71 & 86.85 & 88.62 & 89.54 & 85.22 & 88.47 & 90.56 & 91.09 & 87.48 & (88.31) 88.61 \\
  \hdashline
  \multirow{3}{*}{\bf H3n-Par + mBERT$^\ddag$} & F$_1$ & 87.92 & \bf 92.82 & \bf 87.54 & \bf 90.66 & 93.22 & 95.12 & 89.23 & $-$ & \bf 89.59 &  (90.76) $~~~-~~$ \\
  & UAS & 91.69 & 90.07 & 86.86 & 87.07 & 89.46 & 86.00 & 84.87 & $-$ & 87.23 & (87.91) $~~~-~~$ \\
  & LAS & 89.08 & 89.21 & 85.82 & 85.75 & 86.25 & 84.98 & 83.13 & $-$ & 84.69 & (86.11)  $~~~-~~$ \\
  \bottomrule
\end{tabular}
%}

\end{table*}

In our experiments, we use the same hyperparameter settings as \cite{zhou-zhao-2019-head}. 
For the head scorer, we limit the maximum level to 32, spans exceeding depth 32 are truncated to 32, and we use a hidden layer size 250 for our feedforward networks.
For the pre-trained language models (PrLMs), we use BERT-large-cased \cite{devlin-etal-2019-bert} and XLNet-large \cite{yang2019xlnet} for English in PTB,  RoBERTa-large \cite{liu2019roberta}\footnote{https://github.com/brightmart/roberta\_zh} and XLNet-mid\footnote{https://github.com/ymcui/Chinese-XLNet} for Chinese in CTB, and BERT-base-multilingual-cased for all languages of other benchmarks, including those in English and Chinese.
Following standard practice, we use the EVALB \cite{sekine1997evalb} to calculate the Labeled Precision (LP), Labeled Recall (LR), and Labeled F$_1$ (LF$_1$) for constituency parsing; and report Unlabeled Attachment Score (UAS) and Labeled Attachment Score (LAS) without punctuation for dependency parsing.

\begin{table*}[t]
	\centering
	%\resizebox{1.0\linewidth}{!}{
	\caption{Results on the test datasets of UD v2.6. UDPipe \cite{straka-2018-udpipe,straka2019evaluating}, UDify \cite{kondratyuk-straka-2019-75}, STEPS \cite{grunewald2020graph}. $^*$ indicates the result based on monolingual BERT.}
	\label{tab:ud}
	\begin{tabular}{lcccccccccccc}
		\toprule
		& \multicolumn{2}{c}{ar-PADT} & \multicolumn{2}{c}{cs-PDT} & \multicolumn{2}{c}{de-GSD} & \multicolumn{2}{c}{en-EWT} & \multicolumn{2}{c}{fi-TDT} & \multicolumn{2}{c}{hi-HDTB}\\
		& UAS & LAS & UAS & LAS & UAS & LAS & UAS & LAS & UAS & LAS & UAS & LAS\\
		\midrule
		UDPipe  & 87.54 & 82.94 & 93.33 & 91.31 & 85.53 & 81.07 & 89.63 & 86.97 & 89.88 & 87.46 & 94.85 & 91.83 \\
		\quad + mBERT + Flair & \bf 89.01 & \bf 84.62 & 94.43 & \bf 92.56 & 88.11 & 84.06 & 92.50$^*$ & 90.40$^*$ & 91.66 & 89.49 & 95.56 & 92.50 \\
		UDify + mBERT & 87.72 & 82.88 & 94.73 & 92.88 & 87.81 & 83.59 & 90.96 & 88.50 & 86.42 & 82.03 & 95.13 & 91.46 \\
		STEPS + mBERT & 88.72 & 83.98 & 94.42 & 92.72 & 88.48 & 84.19 & 91.69 & 89.21 & 91.15 & 88.70 & 94.92 & 91.45 \\
		\midrule
		\bf HPSG-Par + mBERT & 88.86 & 83.55 & 94.85 & 92.17 & 88.84 & 84.45 & \bf 93.33$^*$ & \bf 91.29$^*$ & \bf 92.43 & \bf 89.77 & \bf 95.80 & \bf 92.97\\
		\bf H3n-Par + mBERT & 88.58 & 83.17 & \bf 94.87 & 92.18 & \bf 89.04 & \bf 84.63 & 92.51 & 89.17 & 92.22 & 89.46 & 95.69 & 92.81 \\
		\bottomrule
		\toprule
		& \multicolumn{2}{c}{it-ISDT} & \multicolumn{2}{c}{ja-GSD} & \multicolumn{2}{c}{ko-Kaist} & \multicolumn{2}{c}{lv-LVTB} & \multicolumn{2}{c}{ru-SynTagRus} & \multicolumn{2}{c}{zh-GSD}\\
		& UAS & LAS & UAS & LAS & UAS & LAS & UAS & LAS & UAS & LAS & UAS & LAS\\
		\midrule
		UDPipe  & 93.49 & 91.54 & 95.06 & 93.73 & 88.42 & 86.48 & 87.20 & 83.35 & 93.80 & 92.32 & 84.64 & 80.50 \\
		\quad + BERT + Flair & 94.97 & 93.38 & \textbf{95.55} & 94.27 & 89.35 & 87.54 & 88.05 & 84.50 & 94.92 & 93.68 & 90.13$^*$  & 86.74$^*$  \\
		UDify + mBERT  & \bf 95.54 & \bf 93.69 & 94.37 & 92.08 & 87.57 & 84.52 & 89.33 & 85.09 & 94.83 & 93.13 & 87.93 & 83.75 \\
		STEPS + mBERT   & 94.66 & 92.94 & 94.11 & 92.26 & 55.47 & 24.51 & 89.14 & 85.29 & 95.09 & 93.76 & 88.08 & 84.99 \\ 
		\midrule
		\bf HPSG-Par + mBERT & 95.32 & 93.46 &  95.23 & \bf 94.29 & \bf 90.44 & \bf  88.86 & \bf 90.46 &  \bf 87.41 & \bf 95.23 & \bf 93.88 & \bf 90.30 & \bf 87.39 \\
		\bf H3n-Par + mBERT & 95.51 & 93.56 & 95.30 & 94.20 & 90.15 & 88.66 & 90.34 & 87.17 & 95.14 & 93.77 & 90.17 & 87.34\\
		\bottomrule
	\end{tabular}
	%}
	
\end{table*}

\subsection{Results and Analysis}

% In Table \ref{tab:ptb_ctb}, we compare our parser with the latest state-of-the-art related works in constituent and dependency parsing on PTB and CTB benchmarks.
In Table \ref{tab:ptb_ctb}, we compare our parser with the state-of-the-art works in constituent and dependency parsing on PTB and CTB benchmarks.
Although our model enjoys the same decoding time complexity as constituent-only parsers like \cite{kitaev-klein-2018-constituency,tian-etal-2020-improving-constituency}, ours can generate two syntactic parse trees simultaneously, and its performance is still very close to that of the current state-of-the-art parsers.
HPSG-Par and H3n-Par outperform the baselines \cite{dozat2016deep, kitaev-klein-2018-constituency}, indicating the effectiveness of jointly learning constituent and dependency parsing using the HPSG structure.
% Presenting a trade-off between performance and efficiency, H3n-Par has slightly lower overall performance compared with our baseline model HPSG-Par, though this is mostly due to the low accuracy of the head scorer.
Presenting a trade-off between performance and efficiency, H3n-Par has slightly lower overall performance compared with our baseline model HPSG-Par.
With the help of the PrLMs, however, this problem is alleviated.%, as PrLMs improve the head scorer's accuracy. 

Comparing the parsers' parsing speed, the decoding speed of the dependency parsers is faster than that of the constituent parsers when excluding PrLMs, demonstrating how dependency parsing is more computationally friendly than constituent parsing.
When using PrLMs, due to deep neural network computations, the execution of the decoding algorithm takes up a comparatively smaller proportion of the total time compared to when not using PrLMs, but there is still a difference in the average decoding speed.
For SAPar, while there are improvements over its base model (BenePar), in performance due to additional span attention, the decoding speed is reduced.
This phenomenon also occurs comparing KMPar to its base, HPSG-Par; though the proposed labeled attention layer brings benefits to the performance, the decoding efficiency is also affected. Moreover, the complexity of this label attention is seriously linked to the number of labels, which makes it difficult to apply to situations where the syntactic label set is big.
The head scorer structure we introduced is very simple and has a computational complexity of $O(n)$, which is much less than the $O(n^2)$ of the span attention of SAPar and the $O(n^2*|\mathcal{L}|)$ (where $|\mathcal{L}|$ is the number of syntactic labels) of the label attention of KMPar.

Although PrLM computations seem to dominate the time cost when comparing different decoding algorithms by average decoding time,  the average time metric is far from sufficient for evaluating the time complexity of an algorithm because when $n$ is relatively small, decoding algorithms with time complexities $O(n^2)$, $O(n^3)$, and $O(n^5)$ are not much different when performed on modern computers that are well optimized for parallel computing. 
%As $n$ continues to grow, however, the differences arise.

Since the model we proposed supports not only the H3n decoding algorithm, but also CKY (for constituent parse trees), MST (for dependency parse trees), and HPSG (for both parse trees), we can use our model to compare these four algorithms fairly and remove irrelevant factors like model structure and implementation.
For this comparison we use the parser trained on PTB based on XLNet-large PrLM and sample 1000 sentences for parsing in three length intervals (0-100, 100-200, 200-300) of the Wikipedia dumps. 
%The total parsing time and speed of the four decoding algorithms is listed in Table \ref{tab:speed}.
Table \ref{tab:speed} compares four parsing algorithms on parsing time, which shows our H3n yields a remarkable speedup compared to the original HPSG parser.
When sentence length is between 1-100, the parsing time costs of CKY, MST, HPSG, and H3n are in the same order of magnitude, and CKY $<$ MST $<$ H3n $<$ HSPG. Compared with HPSG, our H3n algorithm achieves a 1.41 times decoding speedup, which demonstrates the effectiveness of our proposed H3n algorithm in comparison to the original HPSG algorithm.
In addition, HPSG and H3n both parse faster than the two separately trained constituent and dependency parsers in this setup.

As the sentence length increases to 100-200, the HPSG algorithm with $O(n^5)$ decoding time complexity reveals its limitations more than ten times more time than other algorithms. This shows that the HPSG decoding algorithm is extremely expensive in this situation, despite its performance advantages. 
However, for our proposed H3n decoding algorithm, since the time complexity is reduced from the original $O(n^5)$ to $O(n^3)$, the speedup ratio compared to the original HPSG reaches 17.7. 
Since more factors are considered in our decoding algorithm, as the sequence grows longer, the cost of our additional computations are no longer negligible, so compared using CKY or MST used alone, our decoding algorithm is a bit slower, though using our algorithm to parse both constituent and dependency parse trees is still faster than using a combination of CKY and MST.
When the length further increases to 200-300, the original HPSG decoding is no longer feasible, but our H3n can still work very well and has the same order of magnitude as CKY and MST. In comparison to HPSG, the speedup ratio reaches 639.0.

\section{Towards Universal HPSG Parsing}

According to \cite{zhou-zhao-2019-head} and our main experiments, HPSG parsing shows accuracy superiority over constituent- or dependency-only parsing,
%is an effective approach for modeling constituent and dependency structure directly without additional modification to the model structure, 
but it needs to have two parallel constituent and dependency treebanks for training. This limits the scenarios of its application
%for the HPSG-based parser, as there are often only separate constituent or dependency annotations. 
in multilingual cases as there is often only one of the two treebanks available, which motivates us to find a method to convert a constituent or dependency annotation to the HPSG formalism.
For this reason, we studied the application of HPSG parsing when only constituent or dependency annotations are available.

\subsection{Constituent Treebank Only}%Multilingual Evaluation with 

%The scenario where only constituent syntax available is usually simple because we can 
It is a common practice to convert constituent structures into dependency structures using predetermined head rules. 
%In order to adapt to multiple languages, 
For our multilingual case, we used the UD 1.0 head rules as our basis for converting the multilingual constituent syntactic benchmark SPMRL to a dependency structure.
%to train our HPSG parser. 
The results of the 9 total languages are in Table \ref{table:spmrl}. % reported

From constituent parsing results, we see both HSPG-Par and H3n-Par that use HPSG structure outperforms all other works in all languages except for BenePar in Arabic. This verifies the effectiveness of HPSG-based joint learning and decoding.
%obtain better results excepted in Arabic than the strong baseline BenePar, which is a sole constituent parser, verifying that HPSG structure joint learning and effectiveness of decoding.
Besides, HPSG parsing on the structures converted from different head rules only gives minor differences (90.55 vs. 90.76), which to some extent demonstrates the robustness of HPSG parsing.
Comparing the dependency structure of the conversion using different head rules, we found that the dependency structure of the native conversion in the shared task\footnote{The head rule actually used has not been clarified yet before the deadline.}, the model on the data using UD 1.0 head rules obtained higher results than the native conversion except for Arabic. 
This suggests that the difficulty of parsing the dependency tree obtained from different head rules is different. 
In addition, our H3n-Par also has a significant deviation in the constituent scores between the native conversion and the UD 1.0 conversion, which indicates that different head rules conversion will affect the overall HPSG structure, thereby affecting the parsing performance.

\subsection{Dependency Treebank Only}%Multilingual Evaluation with 

%because the information is lost in the conversion of the constituent trees to the dependency trees. Therefore, the conversion from dependency tree to constituent trees 
The problem becomes a little bit trickier when there are only dependency annotations available, whose conversion usually requires machine learning approaches.
%, and the conditions are strictly limited. 
%In order to remove this obstacle, 
Thus, we developed a conversion algorithm based on HFP. 
With this algorithm\footnote{The implementation of this conversion algorithm will be publicly released once the anonymous review ends.}, we converted the annotations of 12 UD languages and conducted comparative experiments. 
The results shown in Table \ref{tab:ud} indicate that our H3n parser trained on the converted structures generally achieves the highest accuracies on 4 language out of 6 and the rest two only have negligible difference. This verifies the effectiveness of HPSG-based structural 
%In the evaluation results, we have obtained the best parsing scores in multiple languages. 
%In the one hand, this also verifies the effectiveness of HPSG-based structural modeling. 
and additionally shows that our conversion algorithm is helpful. 
Since no additional pre-training Flair embedding is used, our results in some languages are slightly weaker than \cite{straka2019evaluating}, but are also very close.
In addition, H3n-Par is also very close to HPSG-Par in performance, we can conclude so far that H3n optimizes the $O(n^5)$ decoding time complexity of HPSG to $O(n^3)$, which enhances the practicality of HPSG-based parsing, and only at the cost of a small performance loss.

\section{Related Work}

Numerous HPSG-based parsers have been developed in the last decade. For example, the \textit{Enju} English and Chinese parser \cite{miyao2004corpus,yu-etal-2010-semi}, the \textit{Alpine} parser for Dutch \cite{van2006last} and the LKB \& PET \cite{callmeier2000pet,copestake2002implementing} for English, German, and Japanese. 
\cite{sagae-etal-2007-hpsg} presented a solution that incorporates the strengths of dependency parsing into deep HPSG parsing by constraining the application of wide-coverage HPSG rules.
\cite{zhou-zhao-2019-head} formulated constituent and dependency structures as a simplified HPSG for the first time and proposed a graph-based model and joint decoding algorithm.% to parse both syntactic structures simultaneously.
\cite{mrini-etal-2020-rethinking} incorporated an additional label attention layer into \cite{zhou-zhao-2019-head}'s HPSG parser.% for parsing performance improvement.
The recent improvements in syntactic parsing models are mainly in terms of parsing accuracy and neural model structure, while parsing speed and decoding algorithm are often overlooked issues. \cite{li2020global} provides a much faster alternative to using MST in dependency parsing.
%proposed a greedy decoding algorithm in dependency parsing, which has more advantages in decoding speed than MST algorithm.
%In HPSG, a promising parsing modeling approach, reducing the expensive decoding complexity is very valuable.

%In addition, there has been a lot of researches involving 
As for the conversion of dependency trees to constituent trees \cite{xia-palmer-2001-converting,hall-nivre-2008-dependency,wang-zong-2010-phrase,simko-etal-2014-empirical,kong-etal-2015-transforming,lee-wang-2016-language,srinivasan-etal-2020-code}, we take a different motivation by that which we seek the pseudo-constituent trees more in line with the HPSG structure and can handle with both projected and non-projected dependency trees.

% (Zhou & Zhao, 2019) Head-driven phrase structure grammar parsing on Penn treebank
% (Shen et al., 2018) Straight to the tree: Constituency parsing with neural syntactic distance

\section{Conclusion}

Composing and leveraging a head-driven phrase-structure like formalism using constituent and dependency treebanks has improved parsing performance. However, current HPSG parsing suffers time complexity as high as $O$($n^5$) in return for such performance. 
In this work, we focus on this efficiency issue of HPSG parsing. 
Our effort proved fruitful as we successfully preserved HPSG parsing's performance while cutting down decoding time complexity from $O(n^5)$ to $O(n^3)$, making HPSG parsing practical for the first time.
To further this practicality, we also explore options to remove HPSG parsing's need for both constituent and dependency annotations and allow for easier parsing in other languages.
We develop a generally effective dependency-to-constituent conversion to allow us to build a multilingual HPSG parser using existing universal dependency treebanks.
Overall, we present a complete work on more effective, more in-depth, and more general HPSG-based parsing to accommodate more human languages and more universal scenarios.

\bibliographystyle{IEEEtran}
\bibliography{reference}

\end{document}